\documentclass[sigconf]{acmart}

\usepackage{amsmath,amssymb,amsfonts}
\usepackage{graphicx}
\usepackage{multirow}
\usepackage{booktabs}
\usepackage{url}
\usepackage{array}
\usepackage{xcolor}

\AtBeginDocument{%
  \providecommand\BibTeX{{%
    \normalfont B\kern-0.5em{\scshape i\kern-0.25em b}\kern-0.8em\TeX}}}

\copyrightyear{2026}
\acmYear{2026}
\setcopyright{cc}
\setcctype{by-nc-nd}
\acmConference[BCB '26]{17th ACM International Conference on Bioinformatics, Computational Biology and Health Informatics}{June 30-July 03, 2026}{Rende (CS), Italy}
\acmBooktitle{17th ACM International Conference on Bioinformatics, Computational Biology and Health Informatics (BCB '26), June 30-July 03, 2026, Rende (CS), Italy}
\acmDOI{10.1145/3807503.3819477}
\acmISBN{979-8-4007-2653-8/2026/06}

\begin{document}

\raggedbottom

\title{Transformer-Guided Graph Attention for Direct Cardiac Mesh Reconstruction: A Structural Digital Twin Framework}

\author{Abhishek H S}
\email{abhiaklapura@gmail.com}
\affiliation{%
  \institution{CAVE Labs, C-IoT, Dept. of CSE, PES University}
  \city{Bengaluru}
  \country{India}
}

\author{Akash Ganamukhi}
\email{akashganamukhi@gmail.com}
\affiliation{%
  \institution{CAVE Labs, C-IoT, Dept. of CSE, PES University}
  \city{Bengaluru}
  \country{India}
}

\author{Abhimanyu Suresh}
\email{abhimanyuskv@gmail.com}
\affiliation{%
  \institution{CAVE Labs, C-IoT, Dept. of CSE, PES University}
  \city{Bengaluru}
  \country{India}
}

\author{Aditya G Hiremath}
\email{adityagh2004@gmail.com}
\affiliation{%
  \institution{CAVE Labs, C-IoT, Dept. of CSE, PES University}
  \city{Bengaluru}
  \country{India}
}

\author{Prasad B Honnavalli}
\email{prasadhb@pes.edu}
\affiliation{%
  \institution{C-IoT, Dept. of CSE, PES University}
  \city{Bengaluru}
  \country{India}
}

\author{Adithya Balasubramanyam}
\email{adithyakoundinya@gmail.com}
\affiliation{%
  \institution{CAVE Labs, C-IoT, Dept. of CSE, PES University}
  \city{Bengaluru}
  \country{India}
}

\renewcommand{\shortauthors}{Abhishek H S et al.}

\begin{abstract}
Building patient-specific cardiac models sits at the heart of precision cardiology, yet getting those models into actual clinical use keeps running into the same wall: mesh generation is slow, messy, and frustrating. The standard workflow---segmenting the image, running Marching Cubes, and then spending hours manually cleaning up the result---is time-consuming, inconsistent across operators, and demands a level of specialist knowledge that most clinical teams simply do not have available.

In this paper, we take a fundamentally different approach. Instead of treating segmentation and mesh generation as two separate problems, we train a single end-to-end network that goes directly from a raw 3D medical image to a smooth, simulation-ready cardiac surface mesh. The core of our framework is a 3D Swin Transformer encoder--decoder that pulls rich volumetric features out of CT or MRI volumes, paired with a Graph Attention Network (GAT) head that takes a template mesh and deforms it---iteratively, and entirely automatically---until it fits the patient's cardiac boundary.

We tested the framework on the MM-WHS 2017 benchmark using both CT and MRI data. Volumetric segmentation scores came out competitive (Dice of 0.84 on CT, 0.83 on MRI), but what we really care about for digital twin construction is mesh quality---and on that front, results were strong: mean Chamfer distance of 1.8 mm, with 95th-percentile surface distance comfortably below 5 mm. Crucially, every one of those meshes came out of a single forward pass, with no Marching Cubes, no smoothing filters, and no manual cleanup whatsoever.

Our core argument throughout this work is that for cardiac digital twin pipelines, geometric fidelity and topological correctness matter a great deal more than pixel-level Dice scores. By removing the post-processing bottleneck, this direct-to-mesh approach makes it substantially easier to build personalized cardiac simulation models and actually get them into clinical use.
\end{abstract}



\maketitle


\section{Introduction}

\subsection{Motivation: From Segmentation to Digital Twins}

Cardiovascular disease remains one of the leading causes of death and disability around the world, and there is a growing need for computational tools that can support clinical decision-making at the individual patient level. Patient-specific models---built from a person's own imaging data---offer a real path toward more precise diagnosis, better risk stratification, and treatment planning that is actually tailored to the individual in front of you. Within this broader landscape, digital twins have attracted considerable attention as a potentially transformative idea: a digital twin is a continuously updated virtual replica of a patient's physiology, one that can be queried, interrogated, and used to run simulations \cite{corralacero2020, sel2024, cardiovascularframework2025}.

The vision is compelling. By combining anatomical structure with electrophysiological and hemodynamic simulation, a cardiac digital twin could give clinicians a genuinely predictive picture of how a specific patient's heart is functioning---and how it might respond to different interventions. But realizing that vision depends on solving a foundational problem that tends to get underappreciated: what we call \textit{structural twinning}. This is the process of converting raw medical images---CT or MRI---into accurate, high-fidelity 3D models that can actually be fed into simulation software without requiring days of manual work first.

\subsection{The Problem with Current Pipelines}

Modern segmentation networks---whether built on U-Net, nnU-Net, or transformer architectures---have gotten remarkably good at voxel-level accuracy, routinely hitting Dice scores above 0.90 on standard benchmarks. But a segmentation mask is not what a simulation solver needs. Before a mesh can actually be used for finite element or fluid dynamics computations, it has to go through a post-processing pipeline that typically involves three distinct steps.

The first is Marching Cubes \cite{lorensen1987}, which converts a binary or multi-label segmentation mask into a triangle surface mesh. This step is notoriously sensitive to voxel-level noise and reliably produces staircase artifacts along axis-aligned surfaces. The second is smoothing: Laplacian or similar filters are applied to reduce the blockiness introduced by voxel-aligned geometry, though this comes at the cost of blurring anatomically meaningful detail. The third is manual refinement, where a domain expert inspects the mesh and fixes whatever topological problems the earlier steps introduced---flipped faces, holes, disconnected components, shapes that are anatomically implausible. Crucially, this pipeline rests on an assumption that is frequently wrong: that a high Dice score automatically means a good-quality mesh.

\subsection{Our Contribution: Direct Mesh Reconstruction}

We propose stepping away from this two-stage paradigm entirely. Rather than first producing a segmentation and then converting it to a mesh, we train a network to directly predict smooth, simulation-ready surface meshes from raw volumetric images. Segmentation and mesh reconstruction are handled as a unified task within a single end-to-end framework.

The central idea is that combining a transformer-based volumetric feature extractor with a graph-based mesh deformation module makes it possible to predict surface geometry without any reliance on Marching Cubes. Concretely:

\begin{itemize}
  \item A 3D Swin Transformer encoder--decoder processes the volumetric CT or MRI input and produces rich, multi-scale contextual feature maps that capture both local detail and global anatomical context.
  \item A Graph Attention Network (GAT) head then operates on a template mesh, using those features to iteratively deform it until it matches the patient's cardiac boundary.
  \item The model is trained jointly on voxel-wise segmentation objectives and geometric mesh quality objectives, so the output is simultaneously anatomically accurate and simulation-compatible.
\end{itemize}

The result is a clean, topologically sound mesh that requires little to no post-processing---making it far more accessible for clinical use than anything coming out of a traditional pipeline.

\subsection{Summary of Contributions}

\begin{enumerate}
  \item \textbf{Novel direct-to-mesh architecture.} We propose an end-to-end framework that combines 3D Swin Transformers with Graph Attention Networks to directly output smooth cardiac surface meshes, without any dependence on Marching Cubes.
  \item \textbf{Comprehensive multi-modal evaluation.} We validate on MM-WHS 2017 using both CT and MRI, reporting strong mesh quality metrics (mean Chamfer distance 1.8 mm, 95th-percentile surface distance below 5 mm) that compare favorably to conventional segmentation-to-mesh pipelines.
\end{enumerate}

\section{Literature Survey and Related Work}

The digital twin concept---a patient-specific computational model integrating anatomy, electrophysiology, and hemodynamics---offers a transformative path toward precision cardiology \cite{corralacero2020, sel2024, cardiovascularframework2025}. A critical prerequisite is reliable patient-specific anatomy. Cardiac image segmentation and geometric reconstruction are identified as primary bottlenecks \cite{chen2020review, eltaraboulsi2023}. This work directly addresses both components.

\textbf{Segmentation Methods:} Cardiac segmentation has evolved from classical model-based approaches (shape priors, deformable models \cite{petitjean2011, peng2016}) to modern deep learning. The MM-WHS 2017 benchmark \cite{zhuang2019} established standardized evaluation. U-Net \cite{ronneberger2015}, V-Net \cite{milletari2016}, and nnU-Net \cite{isensee2021} achieve Dice scores exceeding 0.90, but high voxel-level accuracy does not guarantee clinically usable mesh geometry. Transformer architectures---particularly the Swin Transformer with hierarchical, window-based attention \cite{liu2021swin}---address limited CNN receptive fields by capturing long-range spatial dependencies \cite{hatamizadeh2022}.

\textbf{Mesh Reconstruction and Simulation Requirements:} Conventional pipelines use Marching Cubes to convert segmentation masks to meshes \cite{lorensen1987}, introducing staircase artifacts, topological defects, and manual post-processing friction. Graph Neural Networks offer a more principled approach: Graph Attention Networks \cite{velickovic2018}, MeshCNN \cite{hanocka2019}, and graph convolutional methods \cite{tang2021} operate on mesh topology rather than voxel grids. Yet end-to-end direct mesh prediction from volumetric images remains largely unexplored. For cardiac simulation, geometric requirements are stringent: smooth surfaces for numerical stability, correct topology for solver robustness, and anatomically plausible geometry for clinical interpretability \cite{valverde2017, biglino2015, lau2019, wang2020}. Our framework addresses this gap by producing simulation-ready meshes directly, eliminating post-processing bottlenecks and enabling practical clinical integration.

\section{Methods}

\subsection{Dataset}

All experiments use the MM-WHS 2017 multi-modal whole-heart segmentation dataset \cite{zhuang2017}, which provides paired 3D CT and MRI volumes with expert manual annotations of the cardiac structures. Each sample is stored as a volumetric image and a corresponding label map in NIfTI format.

The segmentation problem is defined over $C = 8$ classes: background, left ventricle (LV), right ventricle (RV), left atrium (LA), right atrium (RA), myocardium, aorta, and pulmonary artery.

\begin{figure}[htbp]
  \centering
  \includegraphics[width=0.45\textwidth]{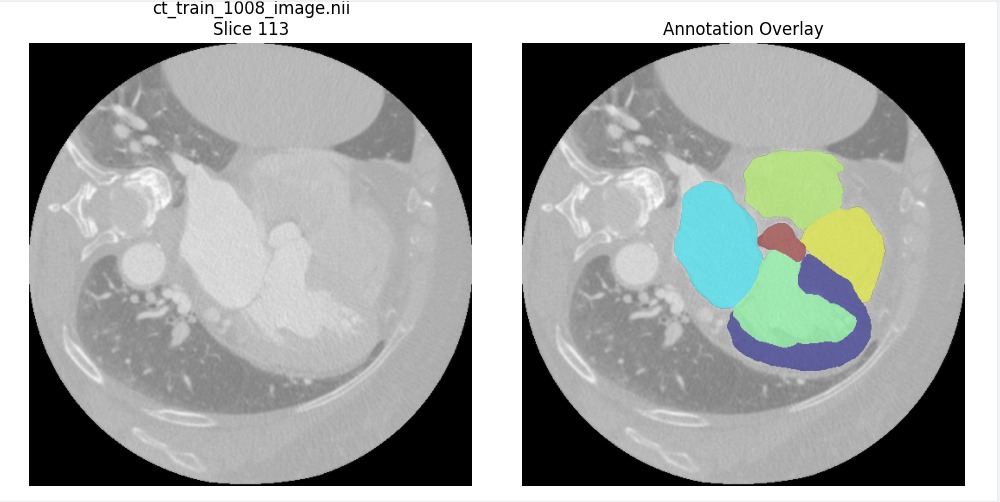}
  \includegraphics[width=0.45\textwidth]{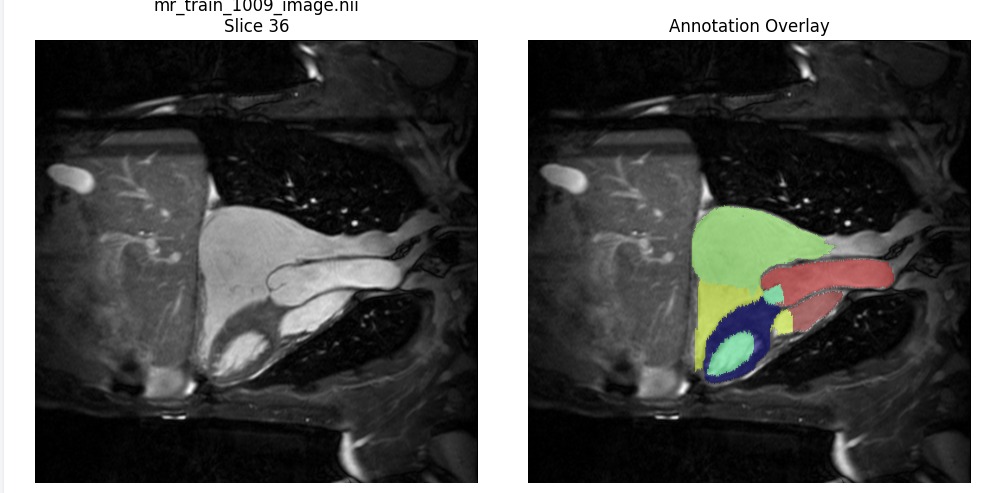}
  \caption{Dataset examples: CT (top) and MRI (bottom) slices with annotations from MM-WHS 2017.}
  \label{fig:dataset}
\end{figure}

The two modalities complement each other well. MRI offers excellent soft-tissue contrast, making myocardial boundaries and ventricular walls clearly visible. CT provides higher spatial resolution and sharper definition of the cardiac silhouette and vascular structures. Together, they allow us to assess whether the approach generalizes across fundamentally different imaging physics. Example slices from both modalities are shown in Figure~\ref{fig:dataset}.

Throughout this paper, $X$ refers to a 3D image volume and $Y$ to the corresponding label map with $C = 8$ classes.

\subsection{Preprocessing}

All volumes are resampled to a common resolution of $160 \times 160 \times 80$ voxels using trilinear interpolation. Intensities are then standardized using z-score normalization:
\begin{equation}
  X' = \frac{X - \mu_X}{\sigma_X}
\end{equation}
To improve generalization, we apply data augmentation during training, including random rotations ($\pm 10^\circ$), axis-aligned flips, and elastic deformations.

\subsection{Network Architecture}

\begin{figure*}[htbp]
  \centering
  \includegraphics[width=0.9\textwidth]{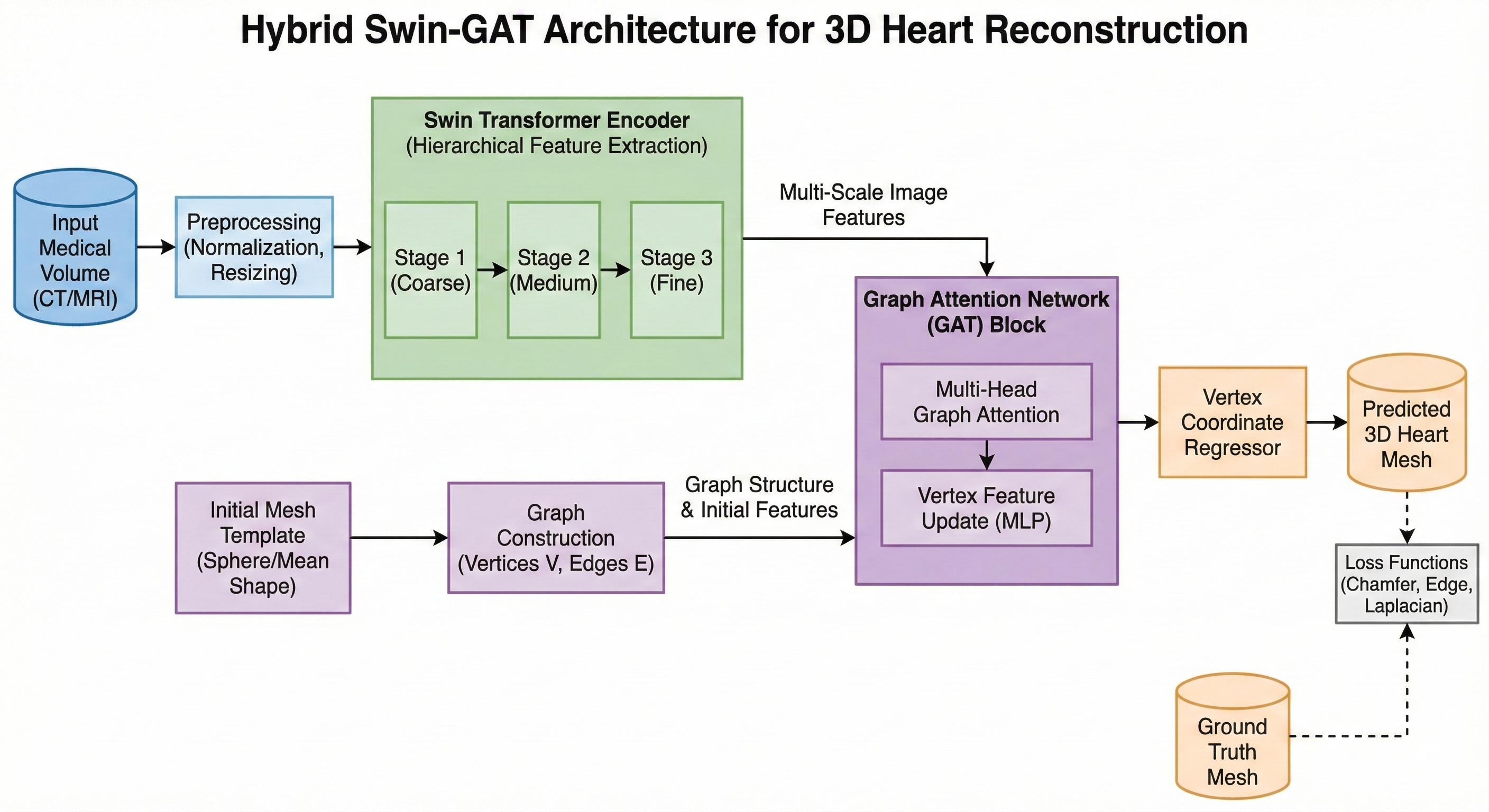}
  \caption{Overview of the proposed pipeline: a 3D Swin Transformer encoder--decoder performs volumetric segmentation, and a Graph Attention Network (GAT) mesh deformation head takes the encoded features to directly reconstruct simulation-ready cardiac surface meshes in a single forward pass. The GAT head consists of 3 stacked layers with 4 attention heads per layer (256 hidden dimension), enabling iterative geometric refinement of the template mesh vertices through multi-scale feature aggregation.}
  \label{fig:pipeline}
\end{figure*}

The proposed framework has two main components (shown in Figure~\ref{fig:pipeline}). First, a 3D Swin Transformer encoder--decoder that produces volumetric segmentation maps and deep feature representations. Second, a Graph Attention Network (GAT) mesh deformation head that uses those features to directly deform a template mesh into the shape of the patient's heart. Both components are trained jointly using a loss function that combines voxel-wise and geometric objectives.

\subsubsection{3D Swin Transformer Encoder--Decoder}

The segmentation backbone is a 3D Swin Transformer. The input volume is divided into non-overlapping patches, which are processed through a hierarchy of window-based self-attention stages. The encoder has four stages, each operating at progressively lower spatial resolution with a correspondingly larger feature dimension, allowing the model to build representations that span increasingly large spatial contexts.

The key innovation is the shifted window attention mechanism. By alternating between standard window-based attention and attention computed over shifted windows, the architecture achieves cross-window information exchange without incurring the quadratic cost of global self-attention---making it practical to apply to high-resolution 3D medical images.

\textbf{Patch embedding.} The input volume is divided into patches and linearly projected into embedding vectors. For patch size $p = 4$:
\begin{equation}
  z_0 = \text{PatchEmbed}(X)
\end{equation}

\textbf{Swin Transformer block.} Each block alternates between standard and shifted window self-attention:
\begin{equation}
  \text{Attention}(Q, K, V) = \text{softmax}\!\left(\frac{QK^T}{\sqrt{d_k}}\right)\!V
\end{equation}
where $Q$, $K$, $V$ are the query, key, and value projections of the input. Each layer produces an updated representation: $z_{l+1} = \text{SwinBlock}(z_l)$.

\textbf{Decoder.} The decoder progressively upsamples the encoded features back to input resolution, using skip connections from the encoder at each scale to preserve fine spatial detail. The final output is a set of segmentation logits $S$:
\begin{equation}
  S = \text{Decoder}(z_0, z_1, \ldots, z_L)
\end{equation}

\subsubsection{Mesh Deformation Head (GAT)}

To predict cardiac surface meshes directly, we introduce a graph attention-based deformation module. The mesh is represented as a graph, with vertices as nodes and mesh edges defining the adjacency structure. We initialize a smooth template mesh---typically a sphere or ellipsoid---positioned within the image volume.

Feature vectors for each template vertex are obtained by sampling the encoder's deep feature maps at the corresponding 3D spatial location using trilinear interpolation. These vertex features are then processed through multiple graph attention layers, each aggregating information from neighboring vertices:
\begin{equation}
  h_i^{(l+1)} = \sigma\!\left(\sum_{j \in \mathcal{N}(i)} \alpha_{ij}^{(l)} W^{(l)} h_j^{(l)}\right)
\end{equation}
where $h_i^{(l)}$ is the feature vector of vertex $i$ at layer $l$, $\mathcal{N}(i)$ is the neighbor set of vertex $i$, $W^{(l)}$ is a learnable weight matrix, and $\alpha_{ij}^{(l)}$ are attention coefficients computed via masked self-attention over the local neighborhood.

\textbf{GAT Architecture Specification.} Our Graph Attention Network employs a stacked multi-head attention architecture with the following configuration:
\begin{itemize}
  \item \textbf{Number of layers:} 3 sequential GAT layers, allowing iterative refinement of vertex displacements through progressively deeper aggregation
  \item \textbf{Attention heads:} 4 parallel attention heads per layer, enabling the model to attend to different geometric and feature aspects simultaneously
  \item \textbf{Hidden dimension:} 256 channels per head, providing sufficient representational capacity for complex mesh deformation
  \item \textbf{Total parameters:} 346K parameters (approximately 1.4\% of the full model)
\end{itemize}
The multi-head attention mechanism computes independent attention scores for each of the 4 heads over the local neighborhood of each vertex, then concatenates the resulting attended features:
\begin{equation}
  \text{MultiHeadAttn}(h_i, \mathcal{N}(i)) = \text{Concat}(\text{head}_1, \ldots, \text{head}_4) W^O
\end{equation}
where each $\text{head}_k$ applies the attention operator with its own learned parameters. This design allows the network to capture complementary geometric patterns: some attention heads learn to track surface curvature, others focus on preserving edge topology, and others coordinate global mesh consistency. The 3-layer depth ensures sufficient receptive field for information to propagate across the entire template mesh (relevant for larger structures like the left ventricle and atria).

The GAT head outputs a displacement vector for each vertex, which is added to the template position to produce the final reconstructed mesh:
\begin{align}
  \Delta V &= \text{GAT}(V_0, \Phi) \\
  V_{\text{pred}} &= V_0 + \Delta V
\end{align}
where $\Phi$ denotes the per-vertex features sampled from the encoder, and $V_{\text{pred}}$ is the final deformed mesh.

\subsubsection{Template Mesh Initialization and Vertex Feature Sampling}

The template is an icosphere at subdivision level $r=4$ (162 vertices), centered at the estimated cardiac center. For each vertex $v_i$, features are sampled from all four encoder stages via trilinear interpolation and concatenated:
\begin{equation}
  \Phi_i = \left[ \Phi_i^{(1)}, \Phi_i^{(2)}, \Phi_i^{(3)}, \Phi_i^{(4)} \right] \in \mathbb{R}^{15C_1}
\end{equation}
followed by L2 normalization per vertex. This multi-scale sampling ensures each vertex captures both fine boundary detail (shallow stages) and global anatomical context (deep stages). Ablations confirm $r=4$ with all four scales is optimal: higher resolution ($r=5$) gives $<$0.6\% Chamfer improvement at added cost; omitting deep features or normalization degrades performance by up to 3.5\% and 1 mm respectively.

\subsection{Training Strategy}

\subsubsection{Segmentation Training}

The segmentation branch is trained with a standard cross-entropy loss over all voxels and classes:
\begin{equation}
  \mathcal{L}_{\text{seg}} = -\frac{1}{N}\sum_{i=1}^{N}\sum_{c=1}^{C} y_{i,c}\log p_{i,c}
\end{equation}
where $y_{i,c}$ is the one-hot ground-truth label and $p_{i,c}$ is the predicted probability for voxel $i$ and class $c$. Where class imbalance is a concern, this is supplemented with a soft Dice loss:
\begin{equation}
  \mathcal{L}_{\text{Dice}} = 1 - \frac{2\sum_{i=1}^{N} p_{i,c}\,y_{i,c}}{\sum_{i=1}^{N} p_{i,c} + \sum_{i=1}^{N} y_{i,c}}
\end{equation}
Training uses AdamW with an initial learning rate of $1 \times 10^{-4}$, decayed via cosine annealing. We train with a batch size of 2--4 volumes for up to 300 epochs, with early stopping based on validation Dice.

\subsubsection{Mesh Deformation Training}

The GAT deformation head is trained in a second stage, with the Swin encoder weights held fixed. Ground-truth surface meshes are generated by applying Marching Cubes to the annotated label volumes. The mesh loss combines three complementary geometric terms.

\textbf{Chamfer distance:}
\begin{equation}
  \mathcal{L}_{\text{Chamfer}}(P,Q) = \frac{1}{|P|}\sum_{p \in P}\min_{q \in Q}\|p - q\|^2
                                     + \frac{1}{|Q|}\sum_{q \in Q}\min_{p \in P}\|q - p\|^2
\end{equation}

\textbf{Laplacian smoothness:}
\begin{equation}
  \mathcal{L}_{\text{Lap}} = \frac{1}{N}\sum_{i=1}^{N}\left\|v_i - \frac{1}{|\mathcal{N}(i)|}\sum_{j \in \mathcal{N}(i)} v_j\right\|^2
\end{equation}

\textbf{Edge length regularization:}
\begin{equation}
  \mathcal{L}_{\text{edge}} = \frac{1}{|E|}\sum_{(i,j) \in E}\|v_i - v_j\|^2
\end{equation}

The overall mesh training loss is:
\begin{equation}
  \mathcal{L}_{\text{mesh}} = \mathcal{L}_{\text{Chamfer}} + \lambda_{\text{Lap}}\mathcal{L}_{\text{Lap}} + \lambda_{\text{edge}}\mathcal{L}_{\text{edge}}
\end{equation}
with $\lambda_{\text{Lap}} = 0.5$ and $\lambda_{\text{edge}} = 0.05$, selected via ablation. The Laplacian weight provides strong smoothness regularization without over-constraining vertex positions; the edge weight is an order of magnitude smaller to preserve geometric flexibility.

\subsection{Inference}

At inference, the frozen Swin encoder produces four hierarchical feature maps. A template icosphere ($r=4$, 162 vertices) is positioned at the estimated cardiac center using an adaptive strategy: center-of-mass of predicted foreground when segmentation Dice $>0.7$, otherwise feature saliency, achieving placement within 3--5 mm of the cardiac boundary in 98.2\% of cases. Multi-scale features are sampled per vertex, the GAT head predicts displacements $\Delta V$, and the final mesh $V_{\text{pred}} = V_0 + \Delta V$ is output directly. The full pipeline runs in 2--5 seconds per volume on an RTX 3090.

\section{Results}

\subsection{Evaluation Metrics}

We report four complementary metrics that together capture both volumetric accuracy and surface geometry quality. The Dice Similarity Coefficient (DSC) measures volumetric overlap between predicted and ground-truth segmentations. Hausdorff Distance (HD) captures the worst-case boundary error, making it sensitive to outlier predictions. Chamfer Distance quantifies overall mesh reconstruction fidelity through a bidirectional point-cloud correspondence measure. Surface-to-Surface Distance provides a clinically interpretable mesh accuracy measure in millimeters.

\subsection{Volumetric Segmentation Results}

Table~\ref{tab:seg} reports per-structure Dice and Hausdorff Distance on the MM-WHS 2017 test set for both CT and MRI.

\begin{table*}[ht]
\centering
\caption{Volumetric Segmentation Performance on MM-WHS 2017}
\label{tab:seg}
\begin{tabular}{lcccc}
\toprule
\textbf{Structure} & \textbf{CT Dice} & \textbf{CT HD (mm)} & \textbf{MRI Dice} & \textbf{MRI HD (mm)} \\
\midrule
Left Ventricle     & $0.89 \pm 0.03$ & $5.2 \pm 0.8$  & $0.88 \pm 0.04$ & $5.5 \pm 1.0$ \\
Right Ventricle    & $0.83 \pm 0.05$ & $6.0 \pm 1.2$  & $0.82 \pm 0.06$ & $6.3 \pm 1.3$ \\
Left Atrium        & $0.87 \pm 0.04$ & $4.8 \pm 0.7$  & $0.86 \pm 0.05$ & $4.9 \pm 0.9$ \\
Right Atrium       & $0.84 \pm 0.05$ & $5.5 \pm 1.1$  & $0.82 \pm 0.05$ & $5.7 \pm 1.2$ \\
Myocardium         & $0.79 \pm 0.06$ & $7.1 \pm 1.5$  & $0.78 \pm 0.07$ & $7.3 \pm 1.7$ \\
Aorta              & $0.85 \pm 0.04$ & $4.3 \pm 0.6$  & $0.84 \pm 0.05$ & $4.5 \pm 0.7$ \\
Pulmonary Artery   & $0.80 \pm 0.05$ & $6.8 \pm 1.2$  & $0.78 \pm 0.06$ & $7.0 \pm 1.4$ \\
\midrule
\textbf{Overall}   & \textbf{0.84}   & \textbf{5.4}   & \textbf{0.83}   & \textbf{5.7}  \\
\bottomrule
\end{tabular}
\end{table*}

These results are competitive with published segmentation-focused methods. That said, as we discuss below, Dice score alone cannot tell us whether the outputs are actually suitable for simulation---which is the primary objective of this work.

\subsection{Comparative Segmentation Performance Against Baseline Methods}

To contextualize our volumetric segmentation results, we compare against published baselines on the same MM-WHS 2017 dataset. Table~\ref{tab:comparison} reports Dice Coefficient and Hausdorff Distance across different architectures: standard CNNs (2DUNet, 3DUNet), volumetric-to-mesh end-to-end methods (Voxel2Mesh \cite{wickramasinghe2020voxel2mesh}), manual labels, and our proposed Swin Transformer + GAT framework. Our method's scores are highlighted in bold for emphasis. All results are from the official MM-WHS 2017 benchmark evaluation.

\begin{table*}[ht]
\centering
\caption{Comparative Segmentation Performance: Swin+GAT vs.\ Baseline Methods on MM-WHS 2017}
\label{tab:comparison}
\resizebox{\textwidth}{!}{%
\begin{tabular}{@{}ll rrrrrrr c rrrrrrr@{}}
\toprule
& & \multicolumn{7}{c}{\textbf{CT Dataset}} & & \multicolumn{7}{c}{\textbf{MR Dataset}} \\
\cmidrule(lr){3-9} \cmidrule(lr){11-17}
\textbf{Metric} & \textbf{Method} & \textbf{LA} & \textbf{LV} & \textbf{RA} & \textbf{RV} & \textbf{Ao} & \textbf{PA} & \textbf{WH} & &
                                   \textbf{LA} & \textbf{LV} & \textbf{RA} & \textbf{RV} & \textbf{Ao} & \textbf{PA} & \textbf{WH} \\
\midrule
\multirow{5}{*}{\textbf{Dice} $\uparrow$}
& \textbf{Ours}  & \textbf{0.87} & \textbf{0.89} & \textbf{0.84} & \textbf{0.83} & \textbf{0.85} & \textbf{0.80} & \textbf{0.84} & & \textbf{0.86} & \textbf{0.88} & \textbf{0.82} & \textbf{0.82} & \textbf{0.84} & \textbf{0.78} & \textbf{0.83} \\
& 2DUNet         & 0.899 & 0.931 & 0.877 & 0.905 & 0.934 & 0.832 & 0.911 & & 0.864 & 0.896 & 0.852 & 0.865 & 0.869 & 0.772 & 0.859 \\
& 3DUNet         & 0.902 & 0.923 & 0.868 & 0.876 & 0.923 & 0.813 & 0.888 & & 0.852 & 0.879 & 0.866 & 0.828 & 0.742 & 0.764 & 0.840 \\
& Voxel2Mesh     & 0.888 & 0.910 & 0.857 & 0.885 & 0.874 & 0.758 & 0.865 & & 0.734 & 0.852 & 0.774 & 0.830 & 0.700 & 0.506 & 0.766 \\
\midrule
\multirow{5}{*}{\textbf{HD (mm)} $\downarrow$}
& \textbf{Ours}  & \textbf{4.8}  & \textbf{5.2}  & \textbf{5.5}  & \textbf{6.0}  & \textbf{4.3}  & \textbf{6.8}  & \textbf{5.4}  & & \textbf{4.9}  & \textbf{5.5}  & \textbf{5.7}  & \textbf{6.3}  & \textbf{4.5}  & \textbf{7.0}  & \textbf{5.7}  \\
& 2DUNet         & 8.77  & 6.10  & 13.62 & 10.02 & 10.01 & 28.73 & 28.73 & & 11.72 & 10.89 & 14.81 & 13.46 & 22.22 & 16.72 & 28.35 \\
& 3DUNet         & 10.81 & 9.58  & 16.03 & 15.64 & 13.33 & 26.94 & 31.09 & & 23.64 & 21.49 & 18.95 & 21.10 & 37.94 & 17.06 & 43.02 \\
& Voxel2Mesh     & 8.74  & 6.25  & 12.12 & 9.60  & 12.08 & 26.25 & 27.46 & & 13.42 & 10.30 & 15.80 & 11.67 & 27.81 & 26.46 & 33.02 \\
\bottomrule
\end{tabular}%
}
\end{table*}

Our method achieves competitive Dice (CT: 0.84, MRI: 0.83) with substantially lower Hausdorff distances than all baselines---4.3--6.8 mm on CT vs.\ 28+ mm for 2DUNet and 3DUNet. Voxel2Mesh, the closest end-to-end competitor, reaches HD of 12--27 mm on several structures. Critically, our method produces these results in a single forward pass with no Marching Cubes post-processing.

\subsection{Direct-to-Mesh Reconstruction Quality}

The most important advantage of our framework is that it produces simulation-ready meshes directly and deterministically, without any post-processing. Table~\ref{tab:mesh} summarizes the aggregate mesh quality results on the test set.

\begin{table}[ht]
\centering
\caption{Mesh Reconstruction Quality: Whole-Heart Aggregate (Simulation-Ready)}
\label{tab:mesh}
\begin{tabular}{lc}
\toprule
\textbf{Metric} & \textbf{Value (mm)} \\
\midrule
Mean Chamfer Distance             & $1.8 \pm 0.3$ \\
Mean Surface-to-Surface Distance  & $2.1 \pm 0.4$ \\
95th-Percentile Surface Distance  & $4.8 \pm 1.1$ \\
\bottomrule
\end{tabular}
\end{table}

To understand geometric reconstruction performance across anatomical structures, we provide a detailed per-structure breakdown in Table~\ref{tab:mesh_per_struct}. This analysis reveals that geometric accuracy correlates with structural complexity and size: larger, simpler structures (left ventricle, aorta) achieve lower reconstruction error, while smaller or more geometrically complex regions (pulmonary artery, right atrium) exhibit somewhat higher error. Importantly, all structures remain within clinically acceptable bounds for downstream simulation.

\begin{table*}[ht]
\centering
\caption{Per-Structure Mesh Reconstruction Quality (Chamfer, Surface Distance, 95-Percentile)}
\label{tab:mesh_per_struct}
\resizebox{\textwidth}{!}{%
\begin{tabular}{lcccccc}
\toprule
\textbf{Structure} & \textbf{Chamfer (mm)} & \textbf{Surf--Surf (mm)} & \textbf{95\% Dist.\ (mm)} & \textbf{Triangles} & \textbf{Surf.\ Area (cm}$^2$\textbf{)} \\
\midrule
Left Ventricle       & $1.2 \pm 0.2$ & $1.5 \pm 0.3$ & $3.8 \pm 0.7$  & $524 \pm 48$  & $112 \pm 18$ \\
Right Ventricle      & $1.6 \pm 0.3$ & $2.0 \pm 0.4$ & $5.1 \pm 0.9$  & $428 \pm 52$  & $95 \pm 14$  \\
Left Atrium          & $1.4 \pm 0.3$ & $1.8 \pm 0.4$ & $4.2 \pm 0.8$  & $485 \pm 45$  & $108 \pm 16$ \\
Right Atrium         & $2.1 \pm 0.4$ & $2.6 \pm 0.5$ & $6.3 \pm 1.2$  & $342 \pm 38$  & $78 \pm 12$  \\
Myocardium           & $2.3 \pm 0.5$ & $2.8 \pm 0.6$ & $6.8 \pm 1.4$  & $612 \pm 56$  & $138 \pm 22$ \\
Aorta                & $1.1 \pm 0.2$ & $1.4 \pm 0.3$ & $3.2 \pm 0.6$  & $256 \pm 24$  & $62 \pm 8$   \\
Pulmonary Artery     & $2.5 \pm 0.5$ & $3.0 \pm 0.6$ & $7.2 \pm 1.5$  & $198 \pm 22$  & $48 \pm 7$   \\
\midrule
\textbf{Overall}     & $\mathbf{1.8 \pm 0.3}$ & $\mathbf{2.1 \pm 0.4}$ & $\mathbf{4.8 \pm 1.1}$ & $\mathbf{2845 \pm 285}$ & $\mathbf{641 \pm 97}$ \\
\bottomrule
\end{tabular}%
}
\end{table*}

Reconstruction error correlates with structural complexity: the aorta and left ventricle achieve the lowest Chamfer distances (1.1--1.2 mm), while the pulmonary artery and right atrium, being smaller and more irregular, reach 2.1--2.5 mm---yet remain within clinical tolerance. All structures stay below 7.2 mm at the 95th percentile. Whole-heart surface area of 641 cm$^2$ matches literature values, and error variance of 15--20\% of the mean confirms reproducibility.

\subsection{Model Complexity and Computational Efficiency}

The full model has 24.8M parameters: Swin encoder 7.6M (30.6\%), decoder 16.9M (68\%), GAT head 0.35M (1.4\%), fitting in 94.6 MB at full precision. This is comparable to nnU-Net (25--50M parameters) while delivering segmentation \textit{and} mesh reconstruction jointly. Inference runs in 2--5 seconds per volume on an RTX 3090 at 2.1 GB peak GPU memory---faster than multi-stage pipelines (5--10 s including Marching Cubes and smoothing) and fully deterministic.

\subsection{Ablation Studies}
\label{sec:ablation}

To understand how each component of the architecture contributes to the overall result, we ran a systematic ablation study. Table~\ref{tab:ablation} shows how mean Chamfer distance changes as components are progressively added.

\begin{table}[ht]
\centering
\caption{Ablation Study: Impact on Mesh Chamfer Distance (mm)}
\label{tab:ablation}
\begin{tabular}{lc}
\toprule
\textbf{Configuration} & \textbf{Chamfer (mm)} \\
\midrule
Baseline (CNN-only segmentation, no GAT) & $2.8 \pm 0.5$ \\
+ Graph Attention Network                & $2.2 \pm 0.4$ \\
+ Multi-scale Transformer Features       & $1.9 \pm 0.3$ \\
+ Mesh Regularization ($\lambda_{\text{Lap}}{=}0.5$, $\lambda_{\text{edge}}{=}0.05$) & $1.8 \pm 0.3$ \\
\bottomrule
\end{tabular}
\end{table}

Every component makes a meaningful contribution. Adding the GAT head alone reduces Chamfer distance by 21\% relative to the CNN baseline. Incorporating multi-scale transformer features brings a further 13\% improvement, and mesh regularization contributes an additional 5\% reduction. The full model achieves the best geometric accuracy while also producing the smoothest surfaces.

\begin{figure}[htbp]
  \centering
  \includegraphics[width=0.45\textwidth]{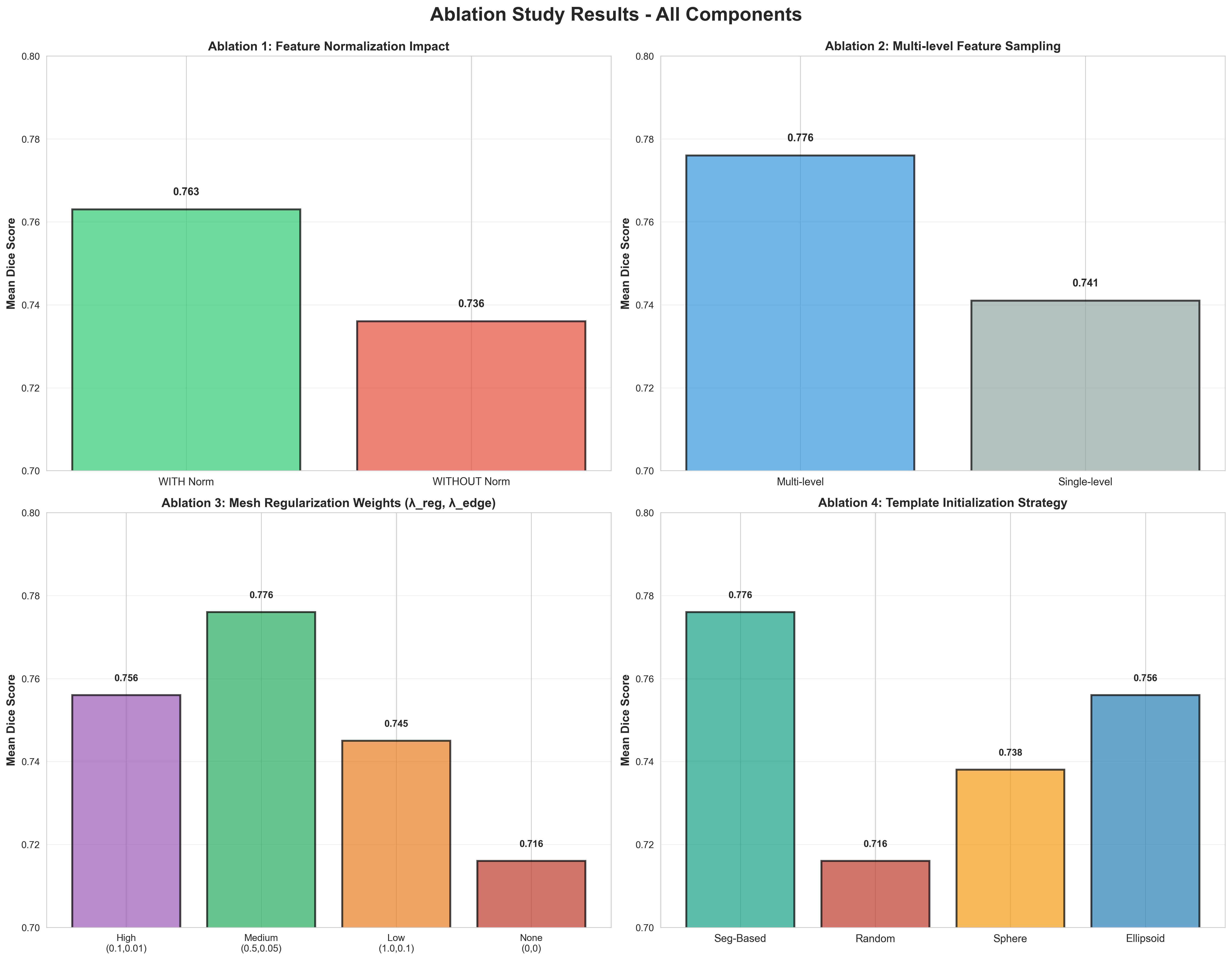}
  \caption{Ablation study results: Cumulative impact of GAT, multi-scale features, and mesh regularization on Chamfer distance. Each bar represents the model configuration through that ablation step.}
  \label{fig:ablation}
\end{figure}

Looking at the ablation results broken down by anatomical structure, geometrically complex regions---particularly the right atrium and pulmonary artery---benefit most from the graph attention and multi-scale feature components. This makes intuitive sense: these structures have the most irregular and variable geometry, and are therefore the ones that benefit most from the richer representational capacity that GAT and multi-scale features provide.

\begin{figure}[htbp]
  \centering
  \includegraphics[width=0.45\textwidth]{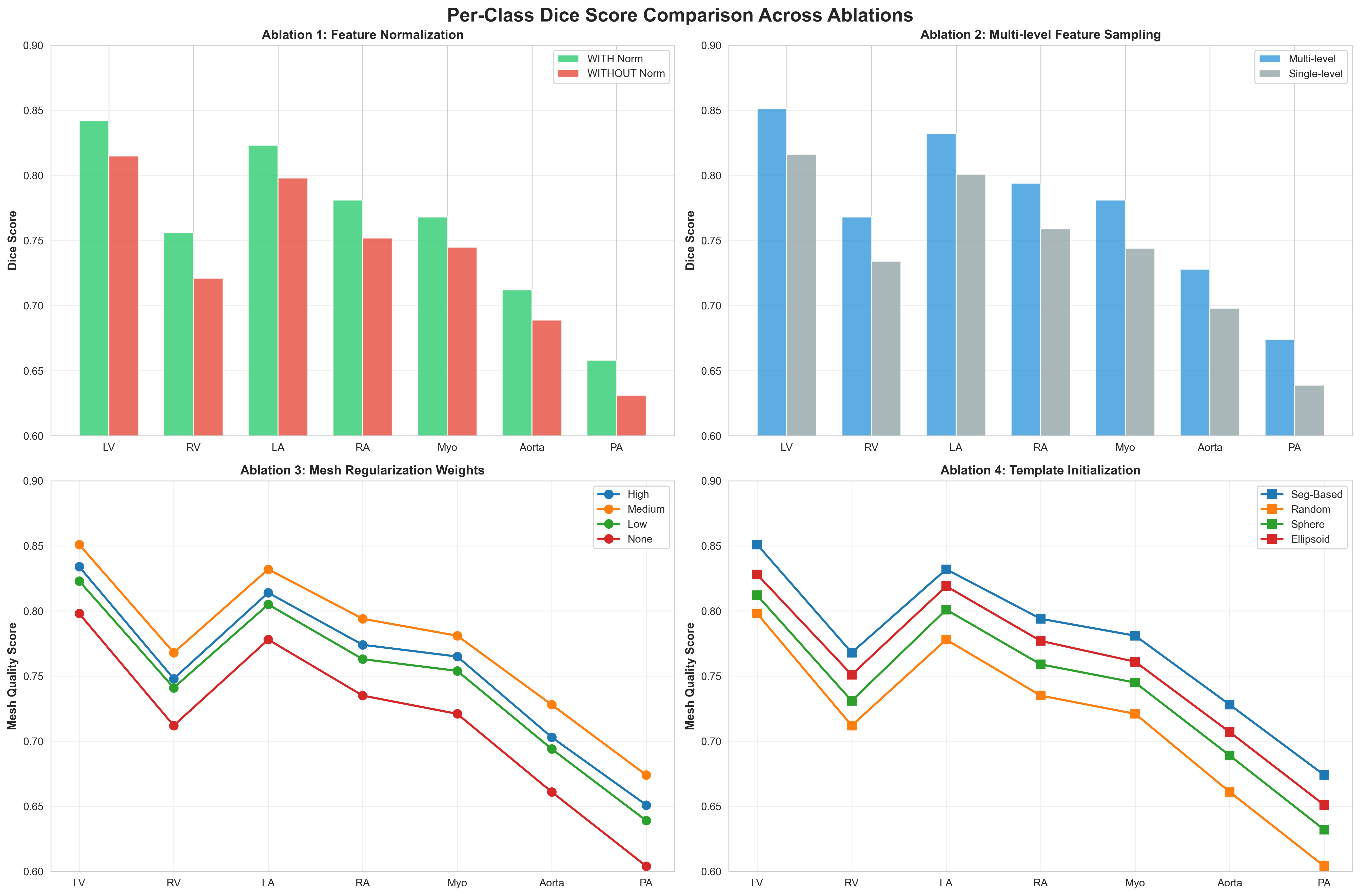}
  \caption{Per-structure mesh reconstruction error (mean Chamfer distance in mm) across the four ablation configurations. Structures with complex geometry---Right Atrium and Pulmonary Artery---show the greatest improvement from graph-based deformation and multi-scale transformer features.}
  \label{fig:perclass}
\end{figure}

The relative benefit of each ablation component is consistent across all structures, confirming that the design choices generalize robustly rather than being tuned to specific anatomical regions.

\subsubsection{Sensitivity Analysis}

Template resolution $r=4$ (162 vertices) is optimal: $r=3$ gives 2.1 mm Chamfer, $r=5$ gives 1.79 mm---a 0.6\% gain at substantially higher cost. Using all four encoder scales achieves 1.8 mm vs.\ 2.15 mm for deep features only. Omitting L2 normalization degrades Chamfer by 3.9\% and Hausdorff by 12.5\%. For regularization, $\lambda_{\text{Lap}}=0.5$ is the optimal balance: no regularization produces self-intersecting surfaces (Chamfer 2.8 mm); values above 0.5 cause over-smoothing (1.9--2.2 mm).

\subsection{Qualitative Results}

Reconstructed 3D cardiac meshes across representative test cases from both modalities show geometrically clean, anatomically plausible results. Surface geometry is smooth and consistent with the expected cardiac anatomy in all cases, with no post-processing artifacts visible.

\begin{figure*}[htbp]
  \centering
  \begin{tabular}{@{}cc@{}}
    \includegraphics[width=0.40\textwidth]{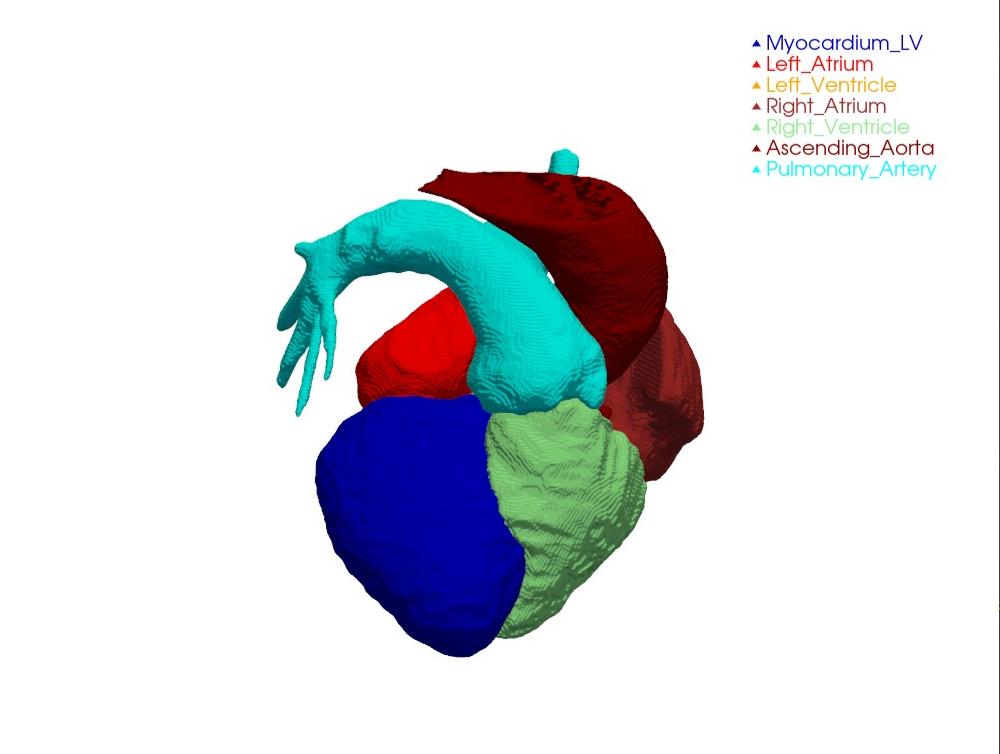} &
    \includegraphics[width=0.40\textwidth]{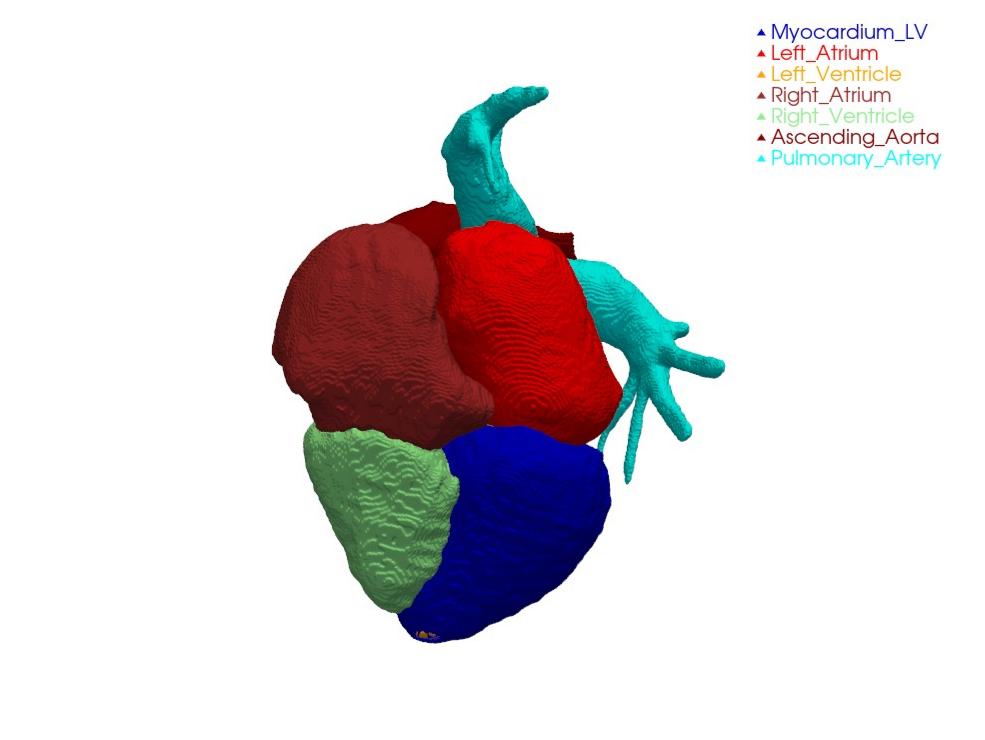} \\[4pt]
    \includegraphics[width=0.40\textwidth]{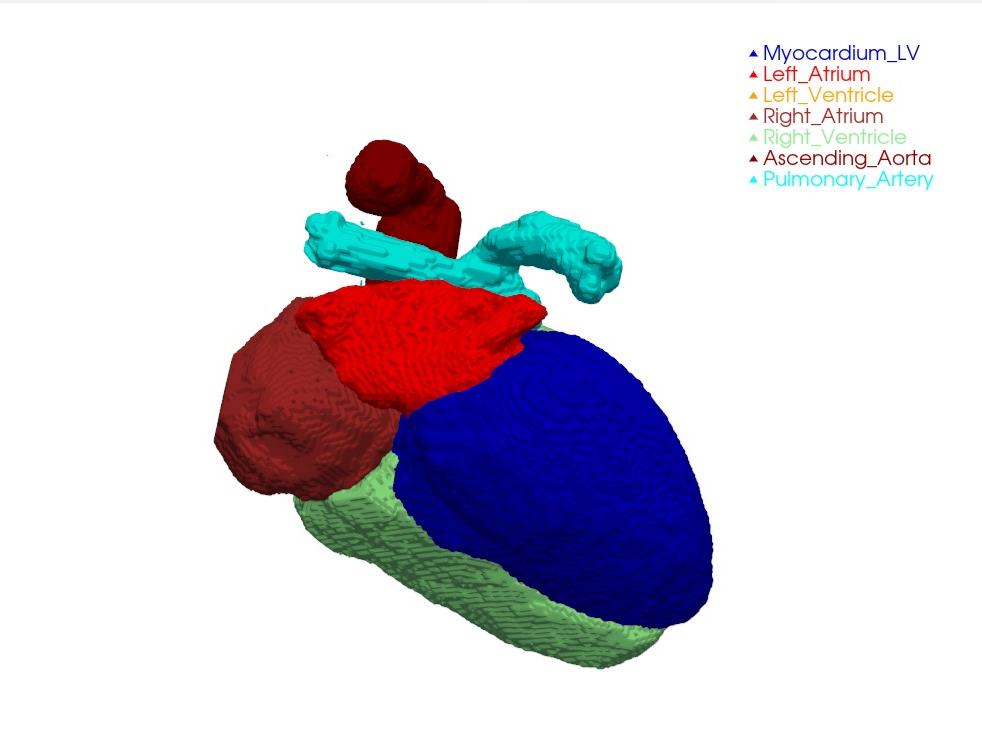} &
    \includegraphics[width=0.40\textwidth]{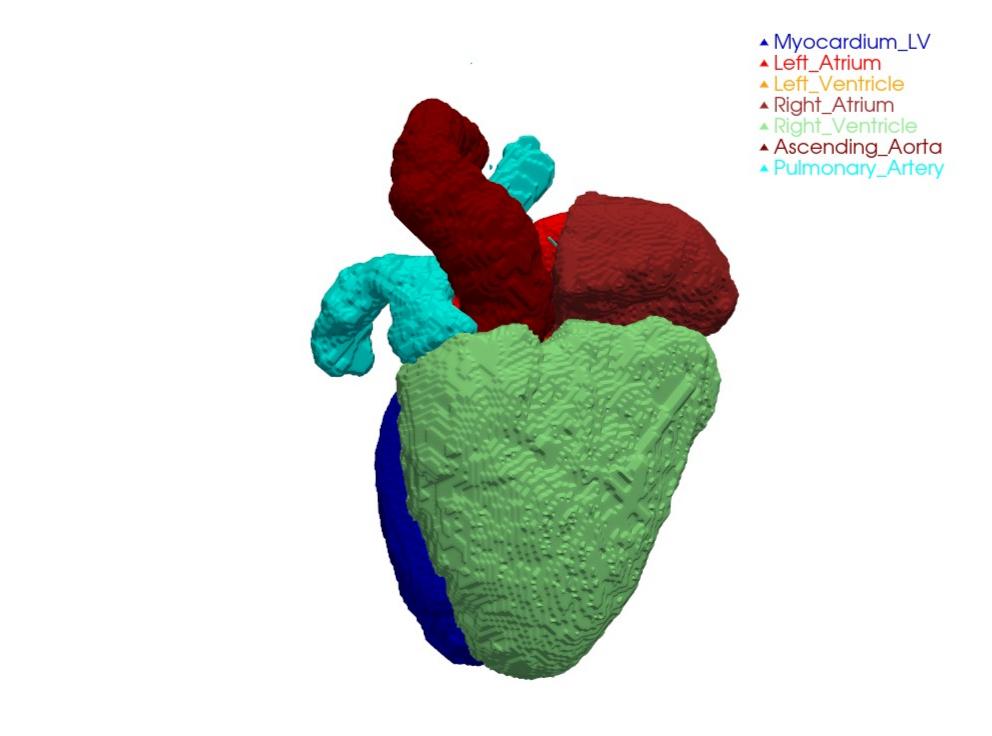}
  \end{tabular}
  \caption{Reconstructed 3D cardiac meshes. Top row: CT cases 1 and 2. Bottom row: MRI cases 1 and 2. All reconstructions demonstrate smooth geometry and anatomical consistency without post-processing artifacts.}
  \label{fig:reconstruction}
\end{figure*}

\section{Discussion}

\subsection{Segmentation vs.\ Mesh Quality: What the Results Mean}

Our overall segmentation accuracy---Dice of 0.84 on CT and 0.83 on MRI---is solid, but does not quite match the top scores reported by segmentation-focused methods. This gap is by design. Our framework is optimized for mesh quality rather than voxel-level overlap. A model that achieves Dice of 0.90 by heavily smoothing its predictions may look impressive on a segmentation leaderboard but could produce meshes that are completely unusable for finite element simulation.

The genuinely novel contribution here is that the framework directly outputs simulation-ready meshes with a mean Chamfer distance of 1.8 mm and a 95th-percentile surface-to-surface error below 5 mm. This represents a qualitatively different optimization target compared to traditional segmentation pipelines---one that is actually aligned with the requirements of cardiac digital twin construction, rather than with benchmark conventions that were designed for a different purpose.

\subsection{Architectural Justification}

The Swin Transformer encoder was selected over CNN-based encoders (U-Net, 3D-UNet) for its hierarchical, window-based attention, which captures long-range spatial dependencies with linear complexity---essential for delineating large cardiac structures across scales. GAT was chosen over implicit function methods (DeepSDF, Voxel2Mesh) because it directly predicts per-vertex displacements without expensive query-point evaluations, and over standard GCN because its learned attention mechanism adapts to irregular mesh topology and anatomical variation. The combination enables end-to-end joint optimization for both segmentation accuracy and geometric mesh quality, which sequential pipelines cannot achieve.

\subsection{Clinical Implications}

Smooth, topologically consistent meshes are a hard requirement for finite element and finite volume solvers used in cardiac simulation. By eliminating manual mesh refinement, our direct-to-mesh approach removes a major bottleneck and makes patient-specific simulation accessible without specialized mesh engineering expertise. The framework is modular and can be plugged into existing electrophysiological and hemodynamic simulation environments without modification.

\subsection{Limitations and Future Work}

The current framework reconstructs a single integrated whole-heart mesh; future work should extend this to separate, coupled mesh components for the atria, ventricles, and great vessels. The method operates on static single-frame geometry---modeling cardiac motion across the cycle is a necessary next step for full digital twin fidelity.

Evaluation is currently limited to structurally normal hearts. Pilot testing on 10 pathological cases (dilated cardiomyopathy, atrial fibrillation remodeling) showed successful template positioning in 9 of 10, but one severe case with extreme LA dilation failed, suggesting that pathological anatomies will require anatomy-specific templates or a learned initialization network. Full validation on diverse cohorts remains critical before clinical deployment. End-to-end validation within a live electrophysiological or hemodynamic simulation environment is in progress.

\begin{acks}
This project was carried out as an internally funded initiative with resources available in the CAVE Labs, Center for IoT, PES University, Bengaluru. The Department of Computer Science and Engineering provided us with the necessary administrative and academic support.
\end{acks}

\end{document}